\documentclass{article}
\usepackage{ijcai26}

\usepackage{times}
\usepackage{soul}
\usepackage{url}
\usepackage[hidelinks]{hyperref}
\usepackage[utf8]{inputenc}
\usepackage[small]{caption}
\usepackage{graphicx}
\usepackage{amsmath}
\usepackage{amsthm}
\usepackage{booktabs}
\usepackage{algorithm}
\usepackage{algorithmic}
\usepackage[switch]{lineno}

\usepackage{xcolor}
\newcommand{\red}[1]{\textcolor{red}{#1}}
\newcommand{\blue}[1]{\textcolor{blue}{#1}}
\usepackage{natbib}      % 引用基础包，支持数字/作者两种格式
\usepackage{amssymb}     % 数学符号，花体、加粗格式
\usepackage{multirow}
\usepackage[table]{xcolor}

\title{H$_{2}$VLR: Heterogeneous Hypergraph Vision-Language Reasoning for Few-Shot Anomaly Detection}

\author{
    Jianghong Huang$^1$                            \and
    Luping Ji$^1$ \footnote{Corresponding author.} \and
    Weiwei Duan$^{1}$                            \and
    Mao ye$^1$                                     \\
\affiliations
    $^1$School of Computer Science and Engineering, University of Electronic Science and Technology of China, Chengdu, China                   \\
\emails
    \{jianghong, weiweiduan\}@std.uestc.edu.cn,
    jiluping@uestc.edu.cn,
    cvlab.uestc@gmail.com
}

\begin{document}

\maketitle

\begin{abstract}
As a classic vision task, anomaly detection has been widely applied in industrial inspection and medical imaging. In this task, data scarcity is often a frequently-faced issue. To solve it, the few-shot anomaly detection (FSAD) scheme is attracting increasing attention. In recent years, beyond traditional visual paradigm,
Vision-Language Model (VLM) has been extensively explored to boost this field.
%%开篇点题结束。
%%归纳目前存在的问题
However, in currently-existing VLM-based FSAD schemes, almost all perform anomaly inference only by pairwise feature matching, ignoring structural dependencies and global consistency.
%high-order
%visual–semantic relations.
%%针对此情况，你做了啥？有什么特色？
To further redound to FSAD via VLM, %Fmitigate data scarcity through pre-trained visual–semantic alignment, but they predominantly rely on independent pairwise matching between visual regions and textual concepts, overlooking the high-order structural dependencies and global consistency that characterize subtle anomalies. 
we propose a \emph{Heterogeneous Hypergraph Vision-Language Reasoning} (\textbf{H$_{2}$VLR}) framework.
It re-formulates the FSAD as a high-order inference problem of visual-semantic relations, by jointly modeling visual regions and semantic concepts in a unified hypergraph. 
%%实验效果如何？
Experimental comparisons verify the effectiveness and advantages of H$_{2}$VLR. It could often achieve state-of-the-art (SOTA) performance on representative industrial and medical benchmarks. Our code will be released upon acceptance.
\end{abstract}

\section{Introduction}
Anomaly detection is a fundamental problem in computer vision, aiming to identify deviations from normal patterns in visual data. It plays a vital role in safety-critical applications such as industrial inspection and medical imaging, where undetected anomalies may lead to severe economic loss or clinical risk \citep{One2normal2024}. In these domains, reliable anomaly detection is essential for ensuring system robustness and decision safety.

\begin{figure}[ht]
    \centering
    \includegraphics[width=0.5\textwidth]{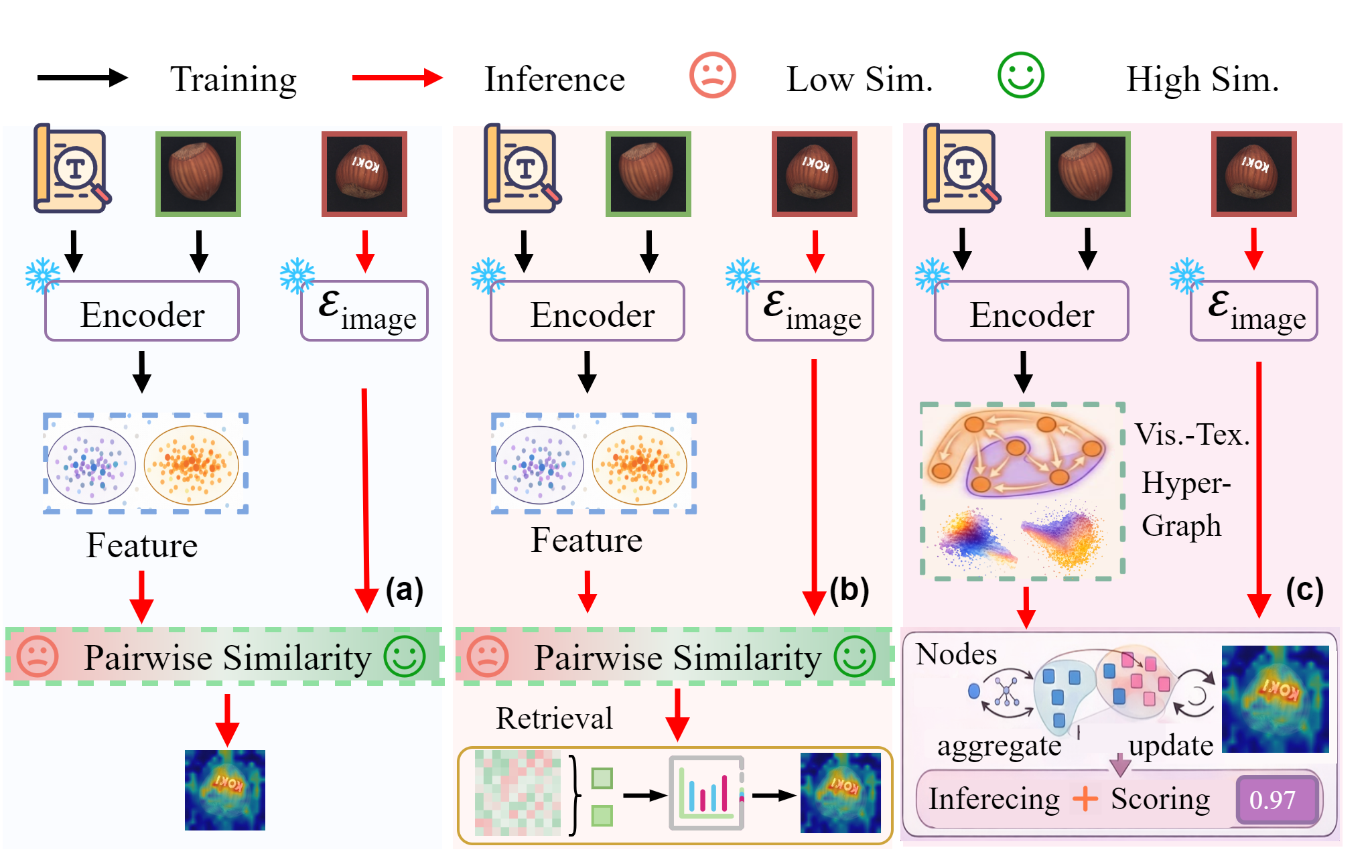}
    \caption{VLM-based FSAD scheme comparison: (a) conventional visual-semantic feature alignment, (b) graph-based feature representation enhancement, (c) our hypergraph scheme, H$_2$VLR.
    %, which performs high-order cross-modal relational inference via a unified hypergraph to capture structural anomalies beyond independent similarity matching.
    }
    \label{fig:3methods}
\end{figure}

In practice, visual anomaly detection is severely constrained by the scarcity of anomalous samples. Anomalies are inherently rare, diverse, and often unpredictable, making it costly and impractical to collect sufficient training data \citep{UniVAD2025}. As a result, conventional data-driven anomaly detection methods struggle to generalize under limited supervision \citep{AdaptCLIP2026}. These challenges have driven an increasing interest in few-shot anomaly detection (FSAD), which performs anomaly detection and localization with limited normal samples \citep{UniNet2025}.

To mitigate data scarcity, recent FSAD approaches increasingly exploit vision-language models (VLMs) such as CLIP \citep{CLIP2021}, which provide strong visual–semantic alignment through large-scale pre-training \citep{FAPrompt2025}. By matching visual features with textual descriptions of normality or abnormality, these methods shift anomaly detection from task-specific training to cross-modal correspondence \citep{WinCLIP2023, PromptAD2024}. As illustrated in Fig.~\ref{fig:3methods}(a), most existing methods adopt prompt-based or patch-level similarity matching, achieving encouraging generalization to unseen categories.

However, such approaches are fundamentally based on independent pairwise matching between isolated visual regions and semantic concepts \citep{AnomalyCLIP2023, AACLIP2025}. This formulation assumes that anomalies can be detected through local similarity deviations, while ignoring the structural dependencies and contextual consistency across regions. As a result, subtle anomalies characterized by coordinated or structural deviations are often overlooked.

Several recent methods attempt to improve visual representations by incorporating graph-based aggregation or memory-driven refinement \citep{KAG-prompt2025}, as shown in Fig.~\ref{fig:3methods}(b). Although these strategies enrich feature representations, their inference stage still relies on independent pairwise matching. Consequently, the reasoning process remains localized and fails to model high-order spatial–semantic relations explicitly. This limitation prevents existing methods from fully capturing the global structural consistency required for reliable anomaly detection under extreme data scarcity.

To overcome these limitations, we propose the H$_{2}$VLR, a hypergraph-based cross-modal reasoning framework that reformulates FSAD as a high-order relational inference problem. Instead of evaluating anomalies through isolated similarity scores, H$_{2}$VLR jointly models visual regions and semantic concepts as heterogeneous nodes within a unified hypergraph, enabling collective reasoning over cross-modal structures. By explicitly capturing high-order dependencies and global contextual consistency, H$_{2}$VLR provides robust anomaly detection and precise localization in low-shot settings. Our main contributions are summarized as follows.

(I) Beyond traditional pairwise matching, we propose the H$_{2}$VLR, a heterogeneous hypergraph Vision-Language reasoning framework for few-shot anomaly detection.
%that goes beyond independent pairwise matching for the detection of anomalies in a few-shots.

(II) We design a group of semantic inducing and hypergraph modeling schemes for H$_{2}$VLR.
%introduce a high-order hypergraph reasoning mechanism that enables detection and localization of structure-aware anomalies under extreme data scarcity.

(III) We design a high-order relation reasoning strategy on visual-semantic feature hypergraph for the effective detection and location inference of image anomaly regions.
%Extensive experiments on industrial and medical benchmarks demonstrate that H$_{2}$VLR consistently outperforms state-of-the-art methods in low-shot settings.

\section{Related Work}
% RW 1
\noindent \textbf{FSAD} aims to characterize normality from extremely limited reference samples and identify deviations at inference time. Early FSAD primarily follows a reconstruction paradigm ($e.g.$, MemAE \citep{MemAE2019}, HTDG \citep{HTDG2021}) to learn normal appearance from few references, while later works shift to feature-matching and memory-augmented paradigms ($e.g.$, PaDiM \citep{PaDiM2021}, PatchCore \citep{PatchCore2022}) that detect anomalies through nearest-neighbor retrieval in pretrained representations.

Recently, many efforts have explored enhanced generalization through transferable learning paradigms. Some meta learning-based methods, such as RegAD \citep{RegAD2022}, enable category-agnostic adaptation through learned comparison mechanisms. Moreover,  some Vision-Language models, such as WinCLIP \citep{WinCLIP2023} and AnomalyCLIP \citep{AnomalyCLIP2023}, introduce semantic supervision through cross-modal alignment. In parallel, some graph-enhanced and diffusion-based models ($e.g.$, DiffAD \citep{DiffAD2023}) incorporate structural cues or generative priors to better model normal distributions.

% RW 2
\noindent \textbf{VLMs} enable semantic-aware FSAD by introducing text as a supervisory signal, alleviating the need for category-specific training. PromptAD \citep{PromptAD2024} leverages textual prompts to condition anomaly scoring, improving cross-category transfer via visual-semantic alignment. Moreover, DictAS \citep{DictAS2025} formulates class-generalizable few-shot anomaly segmentation as dictionary lookup from a few normal references, without target-domain retraining. 

However, existing approaches such as AnomalyCLIP and AdaptCLIP  \citep{AnomalyCLIP2023,AdaptCLIP2026} rely on traditional similarity matching between visual and textual representations, preventing the consistency of abnormal structures, for lacking explicit high-order relational modeling.

% RW 3
\noindent \textbf{Hypergraph learning} generalizes traditional graphs by allowing hyper-edges to connect multiple nodes, effectively capturing high-order correlations beyond simplistic pairwise constraints. For example, the early research \emph{Learning with Hypergraph}~\citep{Zhou2006} establishes a unified spectral learning framework that enables clustering and classification directly on hypergraph structures.

The recent advances in geometric deep learning, $e.g.$, HGNN \citep{HGNN2019}, HyperGCN \citep{HyperGCN2019}, and unified frameworks like UniGNN \citep{UniGNN2021} or AllSet \citep{AllSet2022}, have enabled efficient message-passing for complex structural learning. In computer vision, some paradigms like HgVT \citep{HgVT2025} and HSL \citep{HSL2022} further leverage hypergraphs to model spatial-semantic dependencies. Currently, most hypergraph learning approaches focus on single-modality representations in supervised or closed-set scenarios, with a limited applicability to anomaly detection under few-shot conditions. In particular, how to utilize hypergraph-based reasoning to capture the cross-modal interactions between visual regions and textual prompts still remains largely unexplored.

% 方法部分
\section{Proposed H$_{2}$VLR}

\begin{figure*}[htbp]
    \centering
    \includegraphics[width=0.9\textwidth]{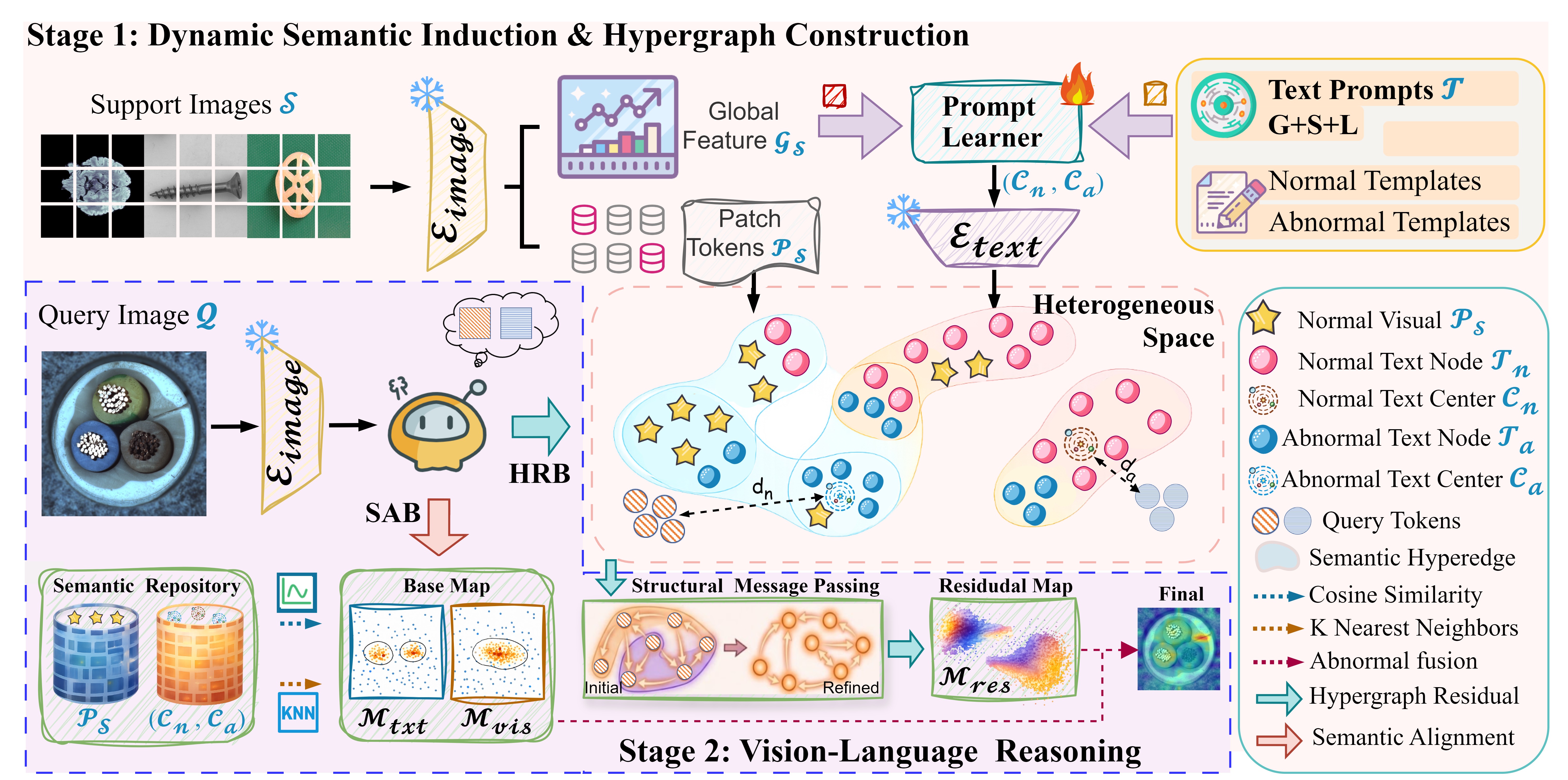}
    \caption{Overview of the proposed H$_2$VLR framework. It  generates the final prediction by fusing a coarse Base Map from the Semantic Alignment Branch (SAB) with a structure-aware Residual Map from the Hypergraph Reasoning Branch (HRB), effectively synergizing the global semantic priors with high-order topological constraints.}
    \label{fig:main}
\end{figure*}

\subsection{Preliminaries}
\label{sec:method_pre}
Different from the standard graphs limited to pairwise relations, we formulate a cross-modal hypergraph $\mathcal{H} = (N, E)$ to capture high-order dependencies among visual regions within a unified vision--language space. We define a heterogeneous node set $N = N_v \cup N_t$, where $N_v = \{p_{q,j}\}$ denotes query visual tokens and $N_t = \{T_k^{(i)}\}$ represents support-conditioned prompt embeddings. Its structural connectivity is governed by the incidence matrix $\mathbf{H} \in \mathbf{R}^{K \times M}$, where $\mathbf{H}_{n,e}=1$ if node $n$ is associated with hyperedge $e$.

\subsection{Method Overview}
\label{sec:method_overall}
As shown in Fig. \ref{fig:main}, we propose the H$_{2}$VLR, a unified cross-modal hypergraph framework for few-shot anomaly detection. Instead of relying on independent image–text similarity estimation, it explicitly models the high-order interactions between adaptive textual semantics and multi-scale visual structures. Firstly, a heterogeneous semantic-guided hypergraph is constructed to embed visual patches and dynamically-generated textual prompts into a shared relational space, where hyperedges encode cross-modal and structural dependencies. 
A hypergraph message passing mechanism is then applied to perform semantic-constrained relational reasoning, enabling structurally meaningful deviations to be emphasized while suppressing unreliable responses. Finally, the refined relational features are incorporated into anomaly inference to yield noise-robust image-level scores and pixel-level precise localization. This formulation provides a compact yet principled pathway from adaptive semantics to high-order structural reasoning and reliable anomaly decision.

\subsection{Semantic Inducing and Hypergraph Modeling}
\label{sec:method_1}
The primary challenge in few-shot industrial inspection lies in the limited expressiveness of generic vision–language priors, where static textual templates often fail to capture category-specific appearance variations and anomaly patterns. To address this issue, we propose Dynamic Semantic Induction (DSI), which conditions textual semantics on the visual statistics of the support images $\mathcal{S}$, yielding task-adaptive textual representations. As shown in the Stage 1 of Fig. \ref{fig:main}, we firstly employ the $\mathcal{E}_{image}$
to extract visual representations from the support samples. Specifically, the support set is summarized by an aggregated global feature  $\mathcal{G}_s \in \mathbb{R}^d$, together with a set of patch-level tokens $\mathcal{P}_s \in \mathbb{R}^{L \times d}$, which preserves local structural information. Rather than directly using these features for similarity matching, $\mathcal{G}_s$ serves as a conditioning signal, adapting the textual prompt space to current task.

To bridge visual and linguistic representations, a learnable mapper projects the global support feature into a sequence of support-aware context tokens, $\mathbf{C}_{\mathcal{S}} = \Psi_{\theta}(\mathcal{G}_s)$. These tokens are injected into a predefined G+S+L (Global–Special–Local) prompt template set ${t}^{0} = {t}_n^0 \cup {t}_a^0$ via an element-wise multiplication $\otimes$, yielding category-adaptive textual embeddings:
\begin{equation}
    \mathcal{T}_k^{(i)} = \mathcal{E}_{text} \left( {t}_k^{0,(i)} \otimes \mathbf{C}_{\mathcal{S}} \right), \quad k \in \{n, a\},
\end{equation}
where $\mathcal{E}_{text}$ denotes the text encoder and $i$ indicates the prompt template numbers within same semantic category. These induced prompts are further aggregated into normal and abnormal semantic centers, $C_k = \text{Avg}(\{ \mathcal{T}_k^{(i)} \})$, forming a semantic repository $\mathcal{R} = \{ \mathcal{T}_n, \mathcal{T}_a, \mathcal{C}_n, \mathcal{C}_a \}$. 

To further regularize the induced semantic space, we impose a margin-based alignment during semantic induction. Let $\text{sim}(\cdot,\cdot)$ denote cosine similarity. For the $j$-th patch token, ${p}_{s,j} \in \mathcal{P}_s$, it requires
\begin{equation}
    \text{sim}({p}_{s,j}, C_n) + \gamma \leq \text{sim}({p}_{s,j}, C_a), \quad \gamma > 0,
\end{equation}
where $\gamma$ denotes a margin hyperparameter, anchoring normal visual patterns to normal semantic center, aiming to enforce it  separating from the abnormal manifold.

% \subsection{Semantic-Guided Hypergraph Modeling}
\label{sec:method_construction}
Building upon the induced semantic repository $\mathcal{R}$, we construct a
cross-modal hypergraph $\mathcal{H}=(N,E)$ to enable high-order relational reasoning, 
beyond traditional patch-level matching. Following the definition in Section \ref{sec:method_pre},
it jointly encodes the structural and semantic dependencies of visual regions in a unified vision-language space.

In detail, we further define a heterogeneous node set $N$ (as defined in Section \ref{sec:method_pre} Preliminaries). In addition, the normal and abnormal semantic centers, $\mathcal{C}_n$ and $\mathcal{C}_a$, are computed by aggregating
the corresponding prompt embeddings, and then serve as global semantic references. While hypergraph is constructed over heterogeneous nodes, hypergraph reasoning is performed exclusively on visual nodes to aggregate semantic–contextual evidence.

The hyperedge set, $E$ is constructed via a dual-guidance strategy. First, structure-preserving hyperedges are formed by connecting each visual patch $p_{q,j}$ to the top-$K$ nearest neighbors in the visual feature space via cosine similarity, enforcing local structural
coherence. Second, semantic-guided hyperedges are instantiated by correlating visual patches with prompt embeddings. Specifically, for each prompt node $\mathcal{T}_k^{(i)}$, we compute cross-modal affinities
${A_{j,k}}^{(i)}=\text{sim}({p}_{q,j},\mathcal{T}_k^{(i)})$ and associate $\mathcal{T}_k^{(i)}$ by the top-$K$ most relevant patches, yielding high-order semantic groupings to link local visual patterns with support-conditioned semantics.

In this case, our hypergraph topology could be represented by an incidence matrix $\mathbf{H}\in\{0,1\}^{|N|\times|E|}$, where $\mathbf{H}_{n,e}=1$ indicates that node $n$ belongs to hyperedge $e$. In implementation, semantic-guided hyperedges are instantiated via a hard top-$K$ membership rule: for a semantic hyperedge indexed by $\mathcal{T}_k^{(i)}$, only the corresponding top-$K$ visual patches are connected. This yields a compact and stable incidence structure, without incurring additional hard-assignment matrices. During inference, it restricts message passing only in the visual subgraph induced by $\mathbf{H}$. Moreover, prompt embeddings and semantic centers influence hypergraph reasoning only through hyperedge construction.

\subsection{High-order Reasoning for Anomaly}
% \subsection{High-order Structural Reasoning}

Based on the heterogeneous incidence structure $\mathbf{H}$ and the prompt-induced edge weights $\boldsymbol{\theta}$, a high-order reasoning module is designed to explicitly infer anomalies as the violations of semantic–structural consistency, rather than traditional patch-level deviations. In few-shot scenarios, local appearance variations are often ambiguous, making a direct similarity-based decision unreliable. Hypergraph reasoning addresses this limitation by propagating anomaly evidence across semantically aligned and structurally coherent regions, enabling context-aware anomaly inference.

Let $\mathbf{X}^{(0)} \in \mathbb{R}^{|N_v| \times d}$ denote the stacked query patch features, where the $j$-th row corresponds to the visual node feature $p_{q,j}$ in $N_v$. Given the incidence matrix $\mathcal{H}$, we adopt a hypergraph convolution operator that follows a node-hyperedge-node aggregation scheme. The node-degree and hyperedge-degree matrices are defined as:
\begin{equation}
    \mathbf{D}_{v}(j, j) = \sum_{e=1}^{|E|} \mathbf{H}_{j, e}, \quad \mathbf{D}_{e}(e, e) = \sum_{j=1}^{|N_{v}|} \mathbf{H}_{j, e}.
\end{equation}
Then, the message passing at layer $\ell$ is formulated by
\begin{equation}
    \mathbf{X}^{(\ell+1)} = \sigma \left( \mathbf{D}_{v}^{-\frac{1}{2}} \mathbf{H} \operatorname{diag}(\boldsymbol{\theta}) \mathbf{D}_{e}^{-1} \mathbf{H}^{\top} \mathbf{X}^{(\ell)} \mathbf{W}^{(\ell)} \right),
\end{equation}
where $\boldsymbol{\theta}$ is the hyperedge weights induced during semantic-guided hyperedge construction, $ \mathbf{W}^{(\ell)}$ is a learnable projection matrix and $\sigma(.)$ denotes a non-linear activation. This operation aggregates information from both structural neighborhoods and semantic-guided groupings, allowing each patch to be contextualized by its high-order relations. By stacking multiple hypergraph reasoning layers, local patch representations are progressively refined through the interactions with semantically consistent and structurally related regions. As a result, anomalous regions will emerge as those that cannot be coherently explained by their hypergraph neighborhoods, rather than merely exhibiting the low similarity to normal semantics. The finally-refined features $\mathbf{X}^{(L)}$ serve as the basis for anomaly saliency estimation and residual fusion.

% \subsection{Hypergraph-Enhanced Anomaly Inference}
Given a query image $\mathcal{Q}$, we extract patch tokens $\mathcal{P_G}$ by the frozen image encoder $\mathcal{E}_{image}$, and compute an init anomaly estimation $\mathcal{M}_{base}$ through the \textbf{S}emantic \textbf{A}lignment \textbf{B}ranch (SAB). Specifically, a text-driven anomaly map $\mathcal{M}_{txt}$ is obtained by the cosine similarity between query patches and the induced semantic repository $\mathcal{R}$. Meanwhile, a visual-driven map $\mathcal{M}_{vis}$ is computed via the KNN matching with the support feature gallery. These two complementary cues are combined together to form the base anomaly map
\begin{equation}
    \mathcal{M}_{base} = \alpha \mathcal{M}_{txt} + (1 - \alpha) \mathcal{M}_{vis} \in [0, 1]^{H \times W},
    \label{eq-anomaly-map}
\end{equation}
which provides a reliable semantic prior while remaining the sensitive to patch-level ambiguity.

To incorporate the structural consistency beyond local patch evidence, we design the \textbf{H}ypergraph \textbf{R}easoning \textbf{B}ranch (HRB). The same query patch features are embedded into a semantic-guided heterogeneous hypergraph, and refined through the $L$ layers of hypergraph reasoning, yielding the final contextualized visual node representations $\mathbf{X}^{(L)} \in \mathbb{R}^{|N_v| \times d}$. An anomaly head is applied to $\mathbf{X}^{(L)}$ to produce node-level probabilities $\mathbf{s}_q \in (0, 1)^{|N_v|}$, which are reshaped and interpolated to the spatial resolution of $\mathcal{M}_{base}$, resulting in a structure-aware anomaly map $\mathcal{M}_{hg} \in (0, 1)^{H \times W}$. Rather than acting as an independent predictor, HRB is designed to refine the base estimation by modeling high-order structural deviations, leading to the hypergraph residual
\begin{equation}
    \mathcal{M}_{res} = \beta (\mathcal{M}_{hg} - \mu) \in [-1, 1] ^{H \times W},
\end{equation}
where $\mu$ denotes the centering bias corresponding to decision boundary, and $\beta$ controls the residual magnitude.

Then, the hypergraph residual is applied as a pixel-wise correction to the base anomaly estimation. For each spatial location $(i,j)$, its correct response is projected into the valid range $[0,1]$ to ensure numerical stability:
\begin{equation}
    \mathcal{M}^*_{i,j} = \min \left\{ 1, \max \left( 0, \mathcal{M}_{base} + \eta\,\mathcal{M}_{res} \right) \right\},
    \label{eq-numerical-stability}
\end{equation}
where $\mathcal{M}^*_{i,j}$ denotes the final anomaly score at pixel $(i,j)$ and $\eta$ controls the contribution of hypergraph reasoning. The final anomaly map $\mathcal{M}^*$ is directly used for pixel-level localization. For image-level detection, $\mathcal{M}^*$ is further summarized by a learnable Soft Histogram Pooling Head, and then fed into a lightweight MLP to produce image-level logits.

\begin{table*}[h]
    \centering
    \renewcommand{\arraystretch}{1.15} % 调行高
    \small
    \begin{tabular}{lccccc}
    \toprule
    Dataset & PromptAD\,(CVPR'24)  & KAG-prompt\,(AAAI'25) & DictAS\,(ICCV'25) & IIPAD\,(ICLR'25) & H$_{2}$VLR \\
     & \citep{PromptAD2024} & \citep{KAG-prompt2025} & \citep{DictAS2025} & \citep{IIPAD}  & (Ours)\\
    \midrule
    MVTec   & (92.17,94.67,95.14) & (91.50,93.98,94.66) & (\blue{93.16},\blue{94.97},\blue{95.93}) & (91.22,93.24,93.91) & (\red{94.98},\red{97.18},\red{97.96}) \\
    VisA    & (80.23,\blue{87.33},\blue{89.09}) & (83.19,86.37,88.63) & (83.82,87.28,88.97) & (\blue{84.50},85.64,86.88) & (\red{88.03},\red{92.08},\red{93.48}) \\
    BTAD    & (90.20,88.58,89.07) & (90.63,91.33,91.59) & (\blue{93.73},\blue{94.19},\blue{94.32}) & (93.50,93.84,94.13) & (\red{94.61},\red{95.19},\red{95.88}) \\
    MPDD    & (72.45,76.74,84.86) & (74.53,79.84,83.22) & (76.37,80.18,85.41) & (\blue{78.64},\blue{81.92},\blue{85.77}) & (\red{79.18},\red{84.22},\red{89.83}) \\
    BeltAD  & (63.64,61.86,70.66) & (62.47,62.85,64.33) & (65.29,65.50,66.39) & (\blue{67.24},\red{68.29},\blue{70.73}) & (\red{67.51},\blue{67.42},\red{74.69}) \\
    \textbf{Average} & (79.74,81.84,85.76) & (80.46,82.87,84.49) & (82.47,84.42,86.20) & (\blue{83.02},\blue{84.59},\blue{86.28}) & (\red{84.86},\red{87.22},\red{90.37}) \\
    \midrule
    BrainMRI & (76.65,76.31,77.33) & (80.02,80.42,82.35) & (\blue{84.11},84.80,85.74) & (82.72,\blue{85.48},\blue{86.09}) & (\red{85.58},\red{85.79},\red{88.11}) \\
    LiverCT  & (63.13,63.04,62.99) & (62.46,64.40,65.09) & (64.97,65.45,65.88) & (\blue{67.05},\blue{67.32},\blue{67.99}) & (\red{69.46},\red{70.03},\red{69.78}) \\
    BUSI     & (\blue{86.21},85.67,83.71) & (81.10,83.41,84.55) & (86.15,\blue{87.29},\blue{89.80}) & (83.75,85.30,88.84) & (\red{90.21},\red{90.13},\red{93.38}) \\
    \textbf{Average} & (75.33,75.01,74.68) & (74.53,76.08,77.33) & (\blue{78.41},79.18,80.47) & (77.84,\blue{79.37},\blue{80.97}) & (\red{81.75},\red{81.98},\red{83.76}) \\
    \bottomrule
    \end{tabular}
    \caption{Image-level anomaly classification performance (AUROC) on industrial (top) and medical (bottom) benchmarks. The best and second-best results are in \red{red} and \blue{blue}. Each triplet $(a, b, c)$ represents the AUROC scores for 1, 2, and 4-shot settings, respectively.}
    \label{table:Image_Res}
\end{table*}

\begin{table*}[h]
    \centering
    \renewcommand{\arraystretch}{1.15} % 调行高
    \small
    \begin{tabular}{lccccc}
    \toprule
    Dataset & PromptAD\,(CVPR'24)  & KAG-prompt\,(AAAI'25) & DictAS\,(ICCV'25) & IIPAD\,(ICLR'25) & H$_{2}$VLR \\
     & \citep{PromptAD2024} & \citep{KAG-prompt2025} & \citep{DictAS2025} & \citep{IIPAD}  & (Ours)\\
    \midrule
    MVTec   & (\blue{95.17},95.77,96.23) & (92.60,94.13,94.65) & (94.98,\blue{95.83},\blue{96.42}) & (94.42,95.21,95.83) & (\red{96.00},\red{97.29},\red{97.81}) \\
    VisA    & (95.10,96.55,97.03) & (95.56,95.74,95.83) & (93.88,94.53,95.54) & (\red{96.79},\blue{97.12},\blue{97.36}) & (\blue{96.46},\red{97.18},\red{97.41}) \\
    BTAD    & (96.18,96.24,96.36) & (96.07,96.13,96.45) & (96.41,\blue{97.05},\blue{97.16}) & (\blue{96.55},96.82,97.14) & (\red{96.83},\red{97.38},\red{97.64}) \\
    MPDD    & (94.73,95.52,96.30) & (95.23,95.79,96.34) & (\blue{96.34},\blue{97.10},\blue{97.59}) & (96.19,96.50,97.34) & (\red{96.68},\red{97.15},\red{97.61}) \\
    BeltAD  & (76.82,75.05,83.91) & (79.88,80.07,84.75) & (79.31,79.75,\blue{85.24}) & (\blue{81.14},\red{81.23},83.83) & (\red{83.02},\blue{80.23},\red{87.93}) \\
    \textbf{Average} & (91.60,91.83,93.97) & (91.87,92.37,93.60) & (92.18,92.85,\blue{94.39}) & (\blue{93.02},\blue{93.38},94.30) & (\red{93.80},\red{93.85},\red{95.68}) \\
    \midrule
    BrainMRI & (92.13,92.39,93.76) & (92.50,93.41,94.96) & (94.88,95.07,95.39) & (\blue{95.22},\blue{95.48},\blue{96.91}) & (\red{96.56},\red{96.73},\red{97.32}) \\
    LiverCT  & (\blue{97.71},\blue{97.56},97.32) & (96.55,96.96,97.14) & (96.87,97.39,97.56) & (97.35,97.54,\blue{97.60}) & (\red{97.79},\red{97.95},\red{97.68}) \\
    BUSI     & (86.06,84.42,85.28) & (88.02,88.71,89.49) & (\blue{88.14},\blue{89.58},90.05) & (87.96,88.87,\blue{91.52}) & (\red{90.06},\red{89.82},\red{92.58}) \\
    \textbf{Average} & (91.97,91.46,92.12) & (92.36,93.03,93.86) &(93.30,\blue{94.01},94.33) & (\blue{93.51},93.96,\blue{95.34}) & (\red{94.80},\red{94.83},\red{95.86}) \\
    \bottomrule
    \end{tabular}
    \caption{Pixel-level anomaly classification performance (AUROC) on industrial (top) and medical (bottom) benchmarks. The best and second-best results are in \red{red} and \blue{blue}. Each triplet $(a, b, c)$ represents the AUROC scores for 1, 2, and 4-shot settings, respectively.}
    \label{table:Pixel_Res}
\end{table*}

\subsection{Optimization Objective}
\label{sec:loss_function}
We optimize this end-to-end framework, $i.e.$, H$_2$VLR, using a compound objective function with three terms:
\begin{equation}
    \label{Ltotal}
    \mathcal{L}_{total} = \mathcal{L}_{align} + \lambda_{str}\mathcal{L}_{struct} + \lambda_{seg}\mathcal{L}_{seg}.
\end{equation}

\noindent\textbf{Loss term 1}: base semantic alignment.
To initialize discriminative node features, we employ a composite alignment loss $\mathcal{L}_{align} = \mathcal{L}_{v2t} + \mathcal{L}_{tri} + \mathcal{L}_{eam}$. Specifically, $\mathcal{L}_{v2t}$ and $\mathcal{L}_{tri}$ align visual tokens with semantic centers and regularize the embedding space, respectively. Moreover, following PromptAD \citep{PromptAD2024}, the Explicit Anomaly Margin (EAM) loss strictly enforces the decision boundary:
\begin{equation}
    \label{Leam}
    \mathcal{L}_{eam} = \mathbb{E}_{\mathbf{z}} \left[ \max \left( 0,\, \text{sim}(\mathbf{z}, \mathcal{C}_a) - \text{sim}(\mathbf{z}, \mathcal{C}_n) + \gamma \right) \right],
\end{equation}
where $\mathbf{z}$ denotes the patch features from training batch.

\noindent\textbf{Loss term 2}: hypergraph structural regularization.
A structural constraint is adopted to propagate sparse supervision across the inferred topology. Let $\mathbf{s}_q$ be the node-level anomaly scores. This hypergraph branch is supervised by
\begin{equation}
    \label{Lstruct}
    \mathcal{L}_{str} = \mathcal{L}_{int}(\mathbf{s}_q, \mathbf{Y}_{node}) + \xi\, \mathbf{s}_q^{\top} \mathbf{L}_{\mathcal{H}} \mathbf{s}_q,
\end{equation}
where $\mathcal{L}_{int}$ combines Focal \citep{Focal2020} and BCE losses against the downsampled ground truth $\mathbf{Y}_{node}$, and $\mathbf{L}_{\mathcal{H}}$ is the Laplacian matrix derived from incidence structure. The quadratic term, weighted by coefficient $\xi$, could enforce manifold smoothness, by penalizing the score discrepancies in structurally-correlated patches.

\noindent\textbf{Loss term 3}: semantic-guided segmentation.
The final map $\mathcal{M}^*$ is supervised by Dice \citep{Dice2016} and Focal losses. To suppress background noise, we incorporate a semantic-weighted penalty:
\begin{equation}
    \label{Lseg}
    \mathcal{L}_{seg} = \mathcal{L}_{dice} + \mathcal{L}_{focal} + \mathbb{E} \left[ (1 - \mathbf{Y}) \otimes \mathbf{W}_{sem} \otimes \mathcal{M}^* \right],
\end{equation}
where $\mathbf{W}_{sem} = 1 + (1 - \mathcal{M}_{txt})$ means that it utilizes the text-driven prior $\mathcal{M}_{txt}$ to down-weight the false positives in semantically irrelevant regions.

% 实验部分
\section{Experiments}

\subsection{Settings}
\noindent{\textbf{Datasets.}} We comprehensively evaluate H$_{2}$VLR on 5 industrial benchmarks (MVTec \citep{MVTec}, VisA \citep{VisA}, BTAD \citep{BTAD}, MPDD \citep{MPDD}, self-constructed BeltAD) and 3 medical datasets (BrainMRI \citep{BrainMRI}, LiverCT \citep{LiverAD}, BUSI \citep{BUSI}).

% multiple datasets from industrial and medical domains. For industrial domain, we use MVTec \citep{MVTec}, VisA \citep{VisA}, BTAD \citep{BTAD}, MPDD \citep{MPDD}, and our proposed BeltAD. In medical domain, we have three medical anomaly detection datasets on different organs like brain, liver and breast (BrainMRI \citep{BrainMRI}, LiverCT \citep{LiverAD} and BUSI \citep{BUSI}). 

\noindent{\textbf{Evaluation Metrics.}} Following the common practice in anomaly detection, we report the Area Under the ROC Curve at both image level (I-AUC) and pixel level (P-AUC), to evaluate detection and localization performance, respectively.

\noindent\textbf{Implementation Details.} 
H${_2}$VLR is implemented in PyTorch with a frozen OpenCLIP (ViT-B/16+) backbone \citep{OpenCLIP}. Optimization is performed by SGD for 100 epochs ($lr$ $2\times10^{-3}$, batch size 400). The image resolution for training and evaluation is $240\times240$. Objective weights are configured as $\lambda_{str}$ $0.02$ and $\lambda_{seg}$ $1.0$. The hypergraph is with $K=8$ and $L=2$. All experiments are conducted on a single NVIDIA RTX 4090 GPU.

\begin{figure*}[t]
    \centering    \includegraphics[width=1\linewidth]{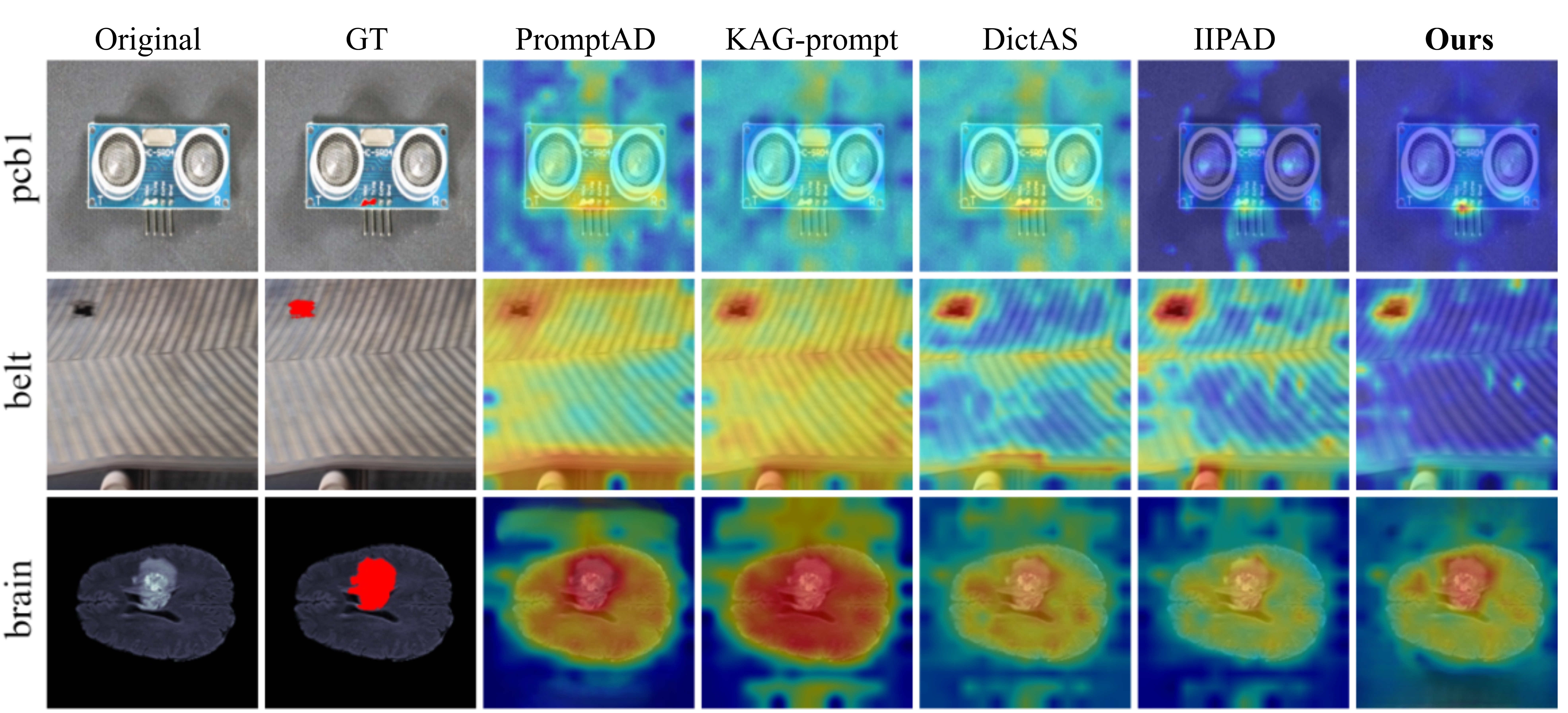}
    \caption{The few-shot anomaly detection  visualization of H$_{2}$VLR, compared to four representative methods.}
    \label{fig:visual_compare}
\end{figure*}

\subsection{Performance Comparisons}

\noindent\textbf{Quantitative Comparison.} We compare H$_{2}$VLR with the four recently-representative methods, including PromptAD \citep{PromptAD2024}, KAG-prompt \citep{KAG-prompt2025}, DictAS \citep{DictAS2025}, and IIPAD \citep{IIPAD}. For fair comparison, all methods are re-trained using official codes under a unified setting. The quantitative results for image-level and pixel-level anomaly detection are summarized in Tab. \ref{table:Image_Res} and Tab. \ref{table:Pixel_Res}, respectively.

%First, H$_{2}$VLR demonstrates consistent superiority and significant performance margins across heterogeneous benchmarks. 

In Tab. \ref{table:Image_Res}, it could be seen that H$_2$VLR could surpass almost other methods in most cases.
For example, under the 4-shot condition, it achieves the highest AUROC on almost all datasets, acquiring the 97.96\% on MVTec, and the 93.38\% on BUSI. 
Similarly, in Tab. \ref{table:Pixel_Res}, it also shows that our H$_2$VLR could often outperform other ones. For example, also under the 4-shot, it still obtains the highest AUROC 97.32\% on BrainMRI. 
Moreover, on both tables, we further observe that H$_2$VLR could obtain better performance on industrial datasets than on medical ones. 
For example, on five industrial datasets (Tab. \ref{table:Image_Res}), it acquires a group of average results (84.86, 87.22, 90.37), clearly superior to the average results (81.75, 81.98, 83.76) obtained on three medical datasets.

Additionally, we further compare H$_{2}$VLR with four representative methods under 8/16-shot and full-samples, as shown in Tab. \ref{tab:full_shot}.
It is easy to see that our 1-shot performance could even surpass the RegAD (8-shot) and AA-CLIP (16-shot) almost on all benchmarks. 
Furthermore, with only 4-shot, our H$_{2}$VLR even outperforms the OneNIP \citep{OneNIP2024} with full-samples on these three industrial datasets. 
For instance, H$_{2}$VLR achieves an I-AUC 97.96\% on MVTec, exceeding the 97.93\% by OneNIP (full-samples). The possible reasons to explain our H$_{2}$VLR superiority could be attributed to the VLM ($i.e.$, CLIP), as well as our hypergraph scheme. 

\begin{table}[h]
    \centering
    \footnotesize
    \resizebox{\columnwidth}{!}{
    \setlength{\tabcolsep}{7pt}
    \begin{tabular}{lcccc}
    \toprule
    Methods & Shots & MVTec & VisA & BTAD \\
    \midrule
    \multirow{2}{*}{\textbf{Ours}} & 1 & 94.98 / 96.00 & 88.03 / 96.46 & \blue{94.61} / 96.83 \\
     & 4 & \red{97.96} / \red{97.81} & \red{93.48} / \blue{97.41} & \red{95.88} / \blue{97.64} \\
    \midrule
    RegAD & 8 & 91.22 / 92.45 & 79.78 / 84.91 & 90.72 / 92.26 \\
    AA-CLIP & 16 & 89.72 / 91.20 & 84.07 / 93.85 & 90.92 / 94.47 \\
    UniAD & full & 96.50 / 95.11 & 90.88 / 96.18 & 92.25 / 92.93 \\
    OneNIP & full & \blue{97.93} / \blue{96.58} & \blue{92.51} / \red{97.49} & 92.79 / \red{98.15} \\
    \bottomrule
    \end{tabular}}
    \caption{Quantitative comparison of I-AUC and P-AUC on industrial benchmarks with many-shot and full-shot methods. }
    \label{tab:full_shot}
\end{table}

\noindent\textbf{Visualization Comparison.}
To intuitively show the anomaly detection and localization capability of H$_{2}$VLR, we present a group of visual results on three representative datasets, in Fig. \ref{fig:visual_compare}. 
By the one-to-one comparison with GT annotations, 
we could obviously observe that H$_{2}$VLR could achieve more precise anomaly detection and location results than any other method, even with less noise, almost on each dataset. These comparisons are exactly consistent with the quantitative results presented in Tab. \ref{table:Image_Res} and Tab. \ref{table:Pixel_Res}, consistently verifying the effectiveness and advantages of our H$_{2}$VLR. 

% more precise anomaly localization by accurately showing the boundaries of anomaly region, while exhibiting superior background noise suppression.

% While baseline methods like PromptAD \citep{PromptAD2024} often generate fragmented or noisy activations, particularly for anomalies with complex morphologies, our approach yields substantially more precise localization. By leveraging high-order relational reasoning to capture global spatial-semantic dependencies, H$_{2}$VLR effectively suppresses background noise and accurately highlights anomalous regions with higher confidence, visually confirming its robustness even under extreme few-shot conditions.

\subsection{Ablation Study}

\noindent\textbf{Module Ablation.} We experimentally verify the effectiveness of individual components in H$_{2}$VLR, including the basic Semantic Alignment Branch (SAB), Dynamic Semantic Induction (DSI), Semantic-Guided Hypergraph Modeling (SGHM), and High-Order Structural Reasoning (HSR). Their experiment results are reported in Tab. \ref{tab:ab_component}.

\begin{table}[h]
    \centering
    \footnotesize
    \resizebox{\columnwidth}{!}{
    \setlength{\tabcolsep}{7pt}
    \begin{tabular}{ccccccc}
        \cmidrule(lr){1-4} \cmidrule(lr){5-7}
        {SAB} & {DSI} & {SGHM} & {HSR}  & {I-AUC} & {P-AUC} & {Mean}\\
        \midrule
        \checkmark & & & & 83.15 & 93.21 & 88.18\\
        \checkmark & \checkmark & & & 85.61 & 94.85 & 90.23 \\
        \checkmark & \checkmark & \checkmark & & 87.12 & 95.83 & 91.46\\
        \checkmark & \checkmark & \checkmark & \checkmark & \red{88.03} & \red{96.46}  & \red{92.32} \\
        \bottomrule
    \end{tabular}}
    \caption{Incremental module ablation with 1-shot setting on VisA. Baseline is only with SAB. The best metric is marked in \red{red}.} 
    \label{tab:ab_component}
\end{table}

It is clearly observed that all modules are effective, and they can always progressively contribute to obvious performance gain. 
Specifically, when DSI is applied, it yields a performance gain of 2.46\% (from 83.15\% to 85.61\%) on I-AUC, validating its effectiveness. 
Moreover, once SGHM and HSR are incrementally adopted, the I-AUC could even further rise from 87.12\% to 88.03\%. 
%Overall, the full model outperforms the SAB baseline by 4.88\% in I-AUC and 3.25\% in P-AUC, confirming the synergy between cross-modal alignment and high-order structural reasoning.

\begin{figure}[h]
    \centering
    \includegraphics[width=\linewidth]{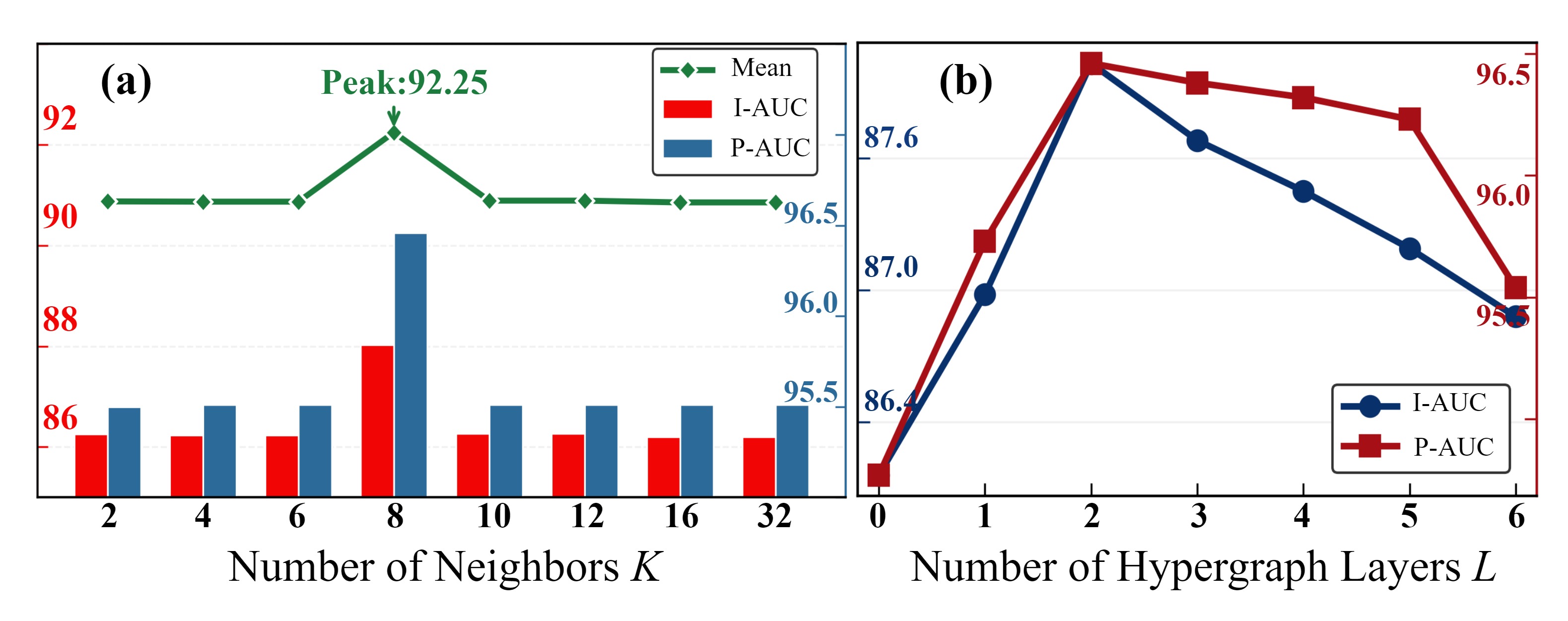}
    \caption{The performance sensitivity to $K$ and $L$, 1-shot, on VisA.}
    \label{fig:hypergraph_para}
\end{figure}

\noindent\textbf{Hypergraph Parameters.} We further evaluate the performance sensitivity to the local topological granularity (by neighbor number $K$) and structural reasoning depth ($L$) in hypergraph.
Their roles are to jointly govern the trade-off between semantic coverage and feature discriminability.

As visualized in Fig.~\ref{fig:hypergraph_para}, we could see two obvious findings. One is that when $K=8$, it will make H$_{2}$VLR to acquire a peak performance of 92.25\% (including I-AUC, P-AUC, and their Mean), in Fig.~\ref{fig:hypergraph_para}(a).
The other is that when $L=2$, H$_{2}$VLR could  obtain its I-AUC and P-AUC performance peaks, in Fig.~\ref{fig:hypergraph_para}(b). 
They imply that the optimal $K$ and $L$ should be set to 8 and 2, respectively.

% increasing the neighbor count $K$ initially enhances performance by incorporating richer geometric contexts, peaking at $K=8$ with a mean accuracy of 92.25\%. However, expanding $K$ beyond this optimal point degrades performance, as overly dense hyperedges introduce irrelevant background noise, leading to semantic contamination. 

% A similar trend is observed with the reasoning depth $L$: while extending message passing from $L=0$ to $L=2$ significantly refines anomaly localization through high-order correlations, further deepening the network triggers the "over-smoothing" phenomenon, rendering node features indistinguishable. Consequently, we empirically adopt $K=8$ and $L=2$ to ensure robust structural modeling.

\begin{figure}[h]
    \centering
    \includegraphics[width=\linewidth]{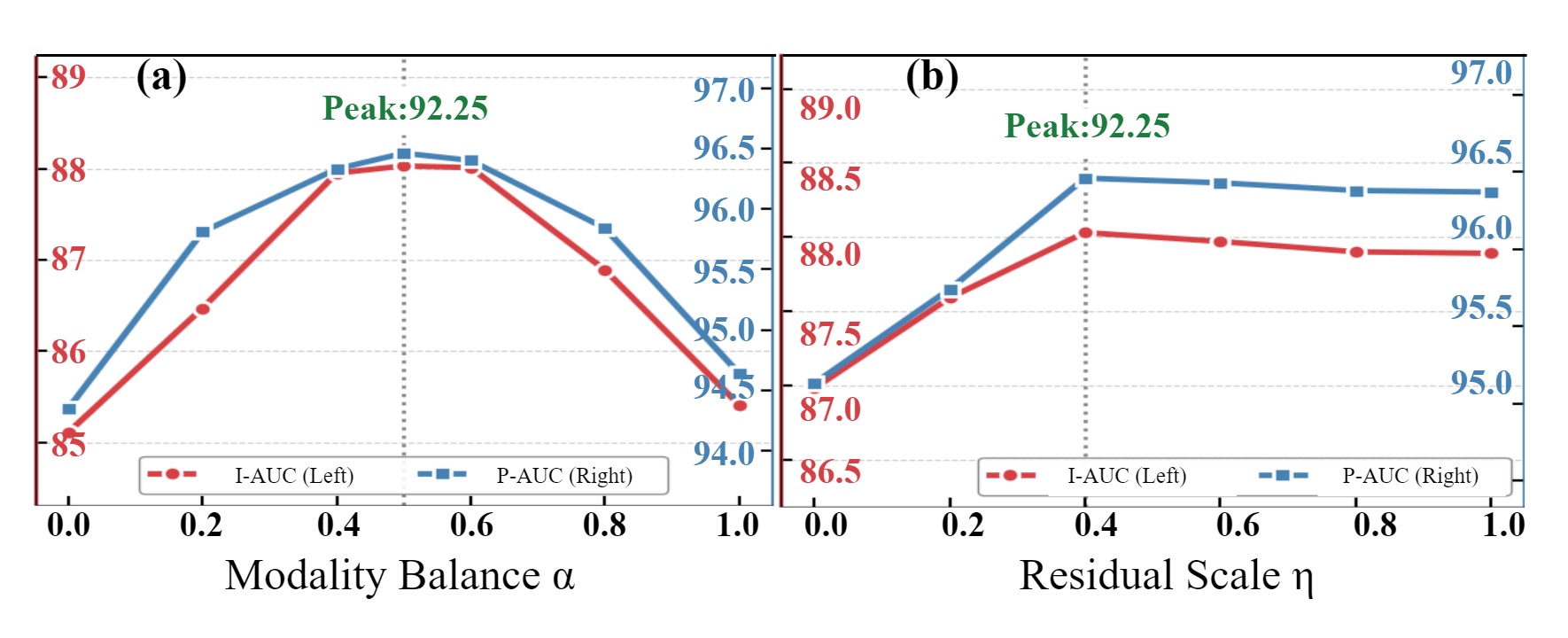}
    \caption{The performance sensitivity to $\alpha$ and $\eta$, 1-shot, on VisA.}
    \label{fig:weight_para}
\end{figure}

\noindent \textbf{Other Primary Parameters.}
Additionally, we also evaluate some other primary hyperparameters, including the $\alpha$ in Eq. (\ref{eq-anomaly-map}) and the $\eta$ in Eq. (\ref{eq-numerical-stability}), as reported in Fig. \ref{fig:weight_para}.

In this figure, it is also easy to see two important observations.
One is that when $\alpha=0.5$, our H$_{2}$VLR could often obtain its peak performance of I-AUC and P-AUC ($i.e.$, their mean value 92.25\%). Moreover, the other is that when $\eta=0.4$, H$_{2}$VLR could reach to its peak performance of I-AUC and P-AUC (meanwhile, their mean value is also 92.25\%). Totally, this group of experiments implies that the optimal $\alpha$ and $\eta$ should be 0.5 and 0.4, respectively.

\begin{table}[h]
    \centering
    \renewcommand{\arraystretch}{1.3} 
    \footnotesize
    \resizebox{\columnwidth}{!}{
    \setlength{\tabcolsep}{3pt}
    \begin{tabular}{l|cc|l|cc}
    \toprule
    $\lambda_{seg}$   & MVTec  & VisA  & $\lambda_{str}$ & MVTec & VisA \\
    \midrule 
    0.5 & 93.37 / 95.14 & 87.10 / 95.25& 0 & 94.82 / 95.91 & 87.97 / 96.41 \\
    0.7 & \blue{94.13} / 95.55 & 87.48 / 95.79 & 0.02 & \red{94.98} / \red{96.00} & \red{88.03} / \red{96.46}\\
     1.0 & \red{94.98} / \red{96.00} & \red{88.03} / \red{96.46}  & 0.04 & \blue{94.87} / \blue{95.96} & \blue{88.01} / \blue{96.44}
    \\
    1.5 & 93.69 / \blue{95.82} & \blue{87.62} / \blue{96.11} & 1.0 & 94.85 / 95.94 & 88.01 / 96.43 \\
    \bottomrule
    \end{tabular}}
    \caption{The impact of $\lambda_{seg}$ and $\lambda_{str}$ on H$_2$VLR performance.}
    \label{tab:loss_abla}
\end{table}

\noindent \textbf{Special Loss Parameters.} To evaluate the sensitivity of primary parameters to performance, we specially test the $\lambda_{seg}$ and $\lambda_{str}$ in Eq. (\ref{Ltotal}), as shown in Tab. \ref{tab:loss_abla}. In this table, we still easily observe two important findings.

One is that when $\lambda_{seg}=1.0$, our H$_{2}$VLR could reach to its optimal performance. For example, when $\lambda_{seg}=0.7$, its I-AUC / P-AUC is only 94.13 / 95.55 on MVTec, while they rise to the highest 94.98 / 96.00 when $\lambda_{seg}=1.0$.  
The other is that when $\lambda_{str}=0.02$, H$_{2}$VLR could acquire its optimal performance 88.03 / 96.46 on VisA. This group of experiments means that the best $\lambda_{seg}$ and $\lambda_{str}$ should usually be set to 1.0 and 0.02, respectively.

% we examine the interplay between pixel-wise supervision and structural consistency in Eq.~(\ref{Ltotal}). Fixing $\lambda_{str}$ at $0.02$, $\lambda_{seg}=1.0$ yields the optimal localization-background trade-off; higher weights (e.g., $1.5$) cause pixel-level gradients to dominate, suppressing structural reasoning. Conversely, the performance drop at $\lambda_{str}=0$ confirms the necessity of manifold smoothness for few-shot stability. 

% in Fig.~\ref{fig:weight_para}. The performance peaks at $\alpha=0.5$, validating the synergy between textual and visual priors. Similarly, a moderate correction strength of $\eta=0.4$ significantly boosts the baseline ($\eta=0$), while excessive correction ($\eta \ge 0.8$) introduces noise, confirming that HRB functions best as a robust refinement module.

\noindent \textbf{Parameters and Inference Speed.} In addition, as shown in Tab. \ref{tab:efficiency}, we finally evaluate the computational overhead of our H$_2$VLR and four other representative methods. 

\begin{table}[h]
    \centering
    \renewcommand{\arraystretch}{1.3} % 将行距设置为原来的 1.5 倍
    \resizebox{\columnwidth}{!}{
    \setlength{\tabcolsep}{5pt}
    \begin{tabular}{llclc}
    \toprule
    Methods & Models & Image Size & Params (M) $\downarrow$ & IPS $\uparrow$ \\
    \midrule
    PromptAD & ViT-B-16+240 & $240 \times 240$ & \textbf{208.4 + 0.1} & \textbf{208} \\
    KAG-prompt & ViT-H-14@224 & $224 \times 224$ & 932.0 + 78.7 & 7 \\
    DictAS & ViT-L/14@336 & $336 \times 336$ & 427.6 + 21.4 & 40 \\
    IIPAD & ViT-L/14@336 & $224 \times 224$ & 427.6 + 8.9 & 44 \\
    \midrule
    \rowcolor[HTML]{EBF5FF} \textbf{Ours} & ViT-B-16+240 & $240 \times 240$ & 208.4 + 2.3 & 33 \\
    \bottomrule
    \end{tabular}}
    \caption{Comparison of model parameters and inference speed.}
    \label{tab:efficiency}
\end{table}

By adopting the ViT-B backbone, it is observed that our H$_2$VLR framework maintains a light-weight network structure, only additionally increasing $2.3$M learnable parameters on backbone (208.4M), although more than the parameter increasing quantity (0.1M) of PromptAD. However, in terms of total parameters, its size (210.7M) is far less than those in other methods ($e.g.$, the 436.5M for IIPAD).

Moreover, in terms of detection speed ($i.e.$, IPS, inference images per second), the peformance metric (33 images, adaptive to real-time applications) of H$_2$VLR seems moderate, far superior to the 7 images by KAG-prompt, although far lower than the 208 images by PromptAD. Nevertheless, in terms of anomaly detection performance, our H$_2$VLR is often the SOTA one (Tab. \ref{table:Image_Res} and Tab. \ref{table:Pixel_Res}). This group of comparisons implies that our H$_2$VLR could obtain SOTA detection performance and real-time inference speed, only by an relatively-low learnable parameter scale. 

% KAG-prompt ($932.0$M) and DictAS ($427.6$M). Although the high-order hypergraph reasoning incurs a slight latency increase compared to the prompt-only baseline (PromptAD), our total inference time of $30.84$ms remains well within the real-time threshold for industrial deployment. This demonstrates a superior trade-off, where H$_2$VLR achieves substantial performance gains at a negligible cost in storage and manageable temporal overhead.

% 总结
\section{Conclusions}
In this work, we propose the H$_{2}$VLR, a Heterogeneous hypergraph Vision-Language Reasoning framework for few-shot anomaly detection. Beyond traditional pairwise matching strategies, it could explicitly models the high-order relations of visual-semantic anomaly features on a unified hypergraph. 
This new strategy enables our method to effectively capture the global structural consistency and spatial–semantic dependencies of anomaly features. 
The experimental evaluations on eight industrial or medical benchmarks demonstrate that our H$_{2}$VLR framework with semantic inducing, hypergraph modeling and high-order reasoning is effective.
It could often consistently achieves SOTA detection performance in the real-world few-shot learning scenarios of anomaly detection. In the future, more complex scenario adaptivity ($e.g.$, 3D anomaly) of our method is worth of exploration.
%%从实验看这个方法有啥不足之处？
%across different few-shot regimes, with strong robustness under extreme data scarcity. These results highlight the importance of high-order relational reasoning as a principled paradigm for identifying subtle anomalies. We believe H$_{2}$VLR provides a scalable and generalizable foundation for visual anomaly reasoning, and future work will extend this framework toward open-world and continually evolving anomaly settings.

% \appendix

% \section*{Acknowledgments}

% This work is supported by the
% National Natural Science Foundation of China (NSFC) under Grants 62476049 and 62276048.

% \input{Supplementary}

\bibliographystyle{named}
\bibliography{ijcai26}

@inproceedings{One2normal2024,
    author = {Li, Yiyue and Zhang, Shaoting and Li, Kang and Lao, Qicheng},
    title = {{One-to-normal: anomaly personalization for few-shot anomaly detection}},
    year = {2024},
    booktitle = {Advances in Neural Information Processing Systems (NeurIPS)},
}

@INPROCEEDINGS{UniVAD2025,
  author={Gu, Zhaopeng and Zhu, Bingke and Zhu, Guibo and $et$ $al$},
  booktitle={2025 IEEE/CVF Conference on Computer Vision and Pattern Recognition (CVPR)}, 
  title={{UniVAD}: A Training-free Unified Model for Few-shot Visual Anomaly Detection}, 
  year={2025},
  volume={},
  number={},
  pages={15194-15203},
}

@INPROCEEDINGS{UniNet2025,
  author={Wei, Shun and Jiang, Jielin and Xu, Xiaolong},
  booktitle={2025 IEEE/CVF Conference on Computer Vision and Pattern Recognition (CVPR)}, 
  title={{UniNet}: A Contrastive Learning-guided Unified Framework with Feature Selection for Anomaly Detection}, 
  year={2025},
  volume={},
  number={},
  pages={9994-10003},
}

@inproceedings{AdaptCLIP2026,
  title={{AdaptCLIP}: Adapting CLIP for Universal Visual Anomaly Detection},
  author={Gao, Bin-Bin and Zhou, Yue and Yan, Jiangtao and $et$ $al$},
  booktitle={AAAI},
  year={2026},
}

@inproceedings{FAPrompt2025,
  title={Fine-grained Abnormality Prompt Learning for Zero-shot Anomaly Detection},
  author={Zhu, Jiawen and Ong, Yew-Soon and Shen, Chunhua and Pang, Guansong},
  booktitle={2025 IEEE/CVF International Conference on Computer Vision (ICCV)},
  year={2025},
}

@InProceedings{CLIP2021,
  title = {Learning Transferable Visual Models From Natural Language Supervision},
  author = {Radford, Alec and Kim, Jong Wook and Hallacy, Chris and $et$ $al$},
  booktitle = {Proceedings of the International Conference on Machine Learning (ICML)},
  pages = {8748-8763},
  year = 	{2021},
  volume = {139},
}

@InProceedings{WinCLIP2023,
    author    = {Jeong, Jongheon and Zou, Yang and Kim, Taewan and $et$ $al$},
    title     = {{WinCLIP}: {Z}ero-/few-shot Anomaly Classification and Segmentation},
    booktitle = {2023 IEEE/CVF Conference on Computer Vision and Pattern Recognition (CVPR)},
    year      = {2023},
    pages     = {19606-19616}
}

@INPROCEEDINGS{PromptAD2024,
  author={Li, Xiaofan and Zhang, Zhizhong and Tan, Xin and $et$ $al$},
  booktitle={2024 IEEE/CVF Conference on Computer Vision and Pattern Recognition (CVPR)}, 
  title={{PromptAD}: Learning Prompts with only Normal Samples for Few-Shot Anomaly Detection}, 
  year={2024},
  volume={},
  number={},
  pages={16848-16858},
}

@InProceedings{AACLIP2025,
    author    = {Ma, Wenxin and Zhang, Xu and Yao, Qingsong and $et$ $al$},
    title     = {{AA-CLIP}: Enhancing Zero-Shot Anomaly Detection via Anomaly-Aware {CLIP}},
    booktitle = {2025 IEEE/CVF Conference on Computer Vision and Pattern Recognition (CVPR)},
    year      = {2025},
    pages     = {4744-4754}
}

@inproceedings{AnomalyCLIP2023,
  title={{AnomalyCLIP}: Object-agnostic Prompt Learning for Zero-shot Anomaly Detection},
  author={Zhou, Qihang and Pang, Guansong and Tian, Yu and He, Shibo and Chen, Jiming},
  booktitle={International Conference on Learning Representations (ICLR)},
  year={2023}
}

@inproceedings{KAG-prompt2025,
    author = {Tao, Fenfang and Xie, Guo-Sen and Zhao, Fang and $et$ $al$},
    title = {Kernel-aware graph prompt learning for few-shot anomaly detection},
    year = {2025},
    articleno = {817},
    numpages = {9},
    booktitle = {AAAI},
}

@inproceedings{MemAE2019,
  title={Memorizing Normality to Detect Anomaly: Memory-augmented Deep Autoencoder for Unsupervised Anomaly Detection},
  author={Gong, Dong and Liu, Lingqiao and Le, Vuong and $et$ $al$},
  booktitle={2019 IEEE/CVF International Conference on Computer Vision (ICCV)},
  year={2019},
pages = {1705-1714},
}

@INPROCEEDINGS{HTDG2021,
  author={Sheynin, Shelly and Benaim, Sagie and Wolf, Lior},
  booktitle={2021 IEEE/CVF International Conference on Computer Vision (ICCV)}, 
  title={A Hierarchical Transformation-Discriminating Generative Model for Few Shot Anomaly Detection}, 
  year={2021},
  volume={},
  number={},
  pages={8475-8484},
}

@InProceedings{PaDiM2021,
    author="Defard, Thomas and Setkov, Aleksandr
    and Loesch, Angelique
    and $et$ $al$",
    title={{PaDiM}: {A} Patch Distribution Modeling Framework for Anomaly Detection and Localization},
    booktitle="Pattern Recognition. ICPR International Workshops and Challenges",
    year="2021",
    pages="475-489",
}

@INPROCEEDINGS{PatchCore2022,
  author={Roth, Karsten and Pemula, Latha and Zepeda, Joaquin and $et$ $al$},
  booktitle={2022 IEEE/CVF Conference on Computer Vision and Pattern Recognition (CVPR)}, 
  title={Towards Total Recall in Industrial Anomaly Detection}, 
  year={2022},
  volume={},
  number={},
  pages={14298-14308},
}

@inproceedings{RegAD2022,
  title={Registration based Few-Shot Anomaly Detection},
  author={Huang, Chaoqin and Guan, Haoyan and Jiang, Aofan and $et$ $al$},
  booktitle={European Conference on Computer Vision (ECCV)},
  year={2022},
}

@INPROCEEDINGS{DiffAD2023,
  author={Zhang, Xinyi and Li, Naiqi and Li, Jiawei and $et$ $al$},
  booktitle={2023 IEEE/CVF International Conference on Computer Vision (ICCV)}, 
  title={Unsupervised Surface Anomaly Detection with Diffusion Probabilistic Model}, 
  year={2023},
  volume={},
  number={},
  pages={6759-6768},
}

@InProceedings{DictAS2025,
    author    = {Qu, Zhen and Tao, Xian and Gong, Xinyi and $et$ $al$},
    title     = {{DictAS}: A Framework for Class-Generalizable Few-Shot Anomaly Segmentation via Dictionary Lookup},
    booktitle = {2025 IEEE/CVF International Conference on Computer Vision (ICCV)},
    year      = {2025},
    pages     = {20519-20528}
}

@inproceedings{zhou2006,
    author = {Zhou, Dengyong and Huang, Jiayuan and Sch\"{o}lkopf, Bernhard},
    title = {Learning with hypergraphs: clustering, classification, and embedding},
    year = {2006},
    booktitle = {Advances in Neural Information Processing Systems (NeurIPS)},
    pages = {1601–1608},
}

@inproceedings{HGNN2019,
  title={Hypergraph Neural Networks},
  author={Feng, Yifan and You, Haoxuan and Zhang, Zizhao and Ji, Rongrong and Gao, Yue},
  numpages = {8},
  booktitle = {AAAI},
  year={2018}
}

@incollection{HyperGCN2019,
    title = {{HyperGCN}: A New Method For Training Graph Convolutional Networks on Hypergraphs},
    author = {Yadati, Naganand and Nimishakavi, Madhav and Yadav, Prateek and $et$ $al$},
    booktitle = {Advances in Neural Information Processing Systems (NeurIPS)},
    pages = {1509--1520},
    year = {2019},
}

@inproceedings{UniGNN2021,
  title     = {{UniGNN}: {A} Unified Framework for Graph and Hypergraph Neural Networks},
  author    = {Huang, Jing and Yang, Jie},
  booktitle = {Proceedings of the Thirtieth International Joint Conference on Artificial Intelligence, {IJCAI-21}},
  year      = {2021},
  pages     = {2563--2569},
}

@inproceedings{AllSet2022,
    title={{You are AllSet}: {A} Multiset Function Framework for Hypergraph Neural Networks},
    author={Eli Chien and Chao Pan and Jianhao Peng and Olgica Milenkovic},
    booktitle={International Conference on Learning Representations (ICLR)},
    year={2022},
}

@InProceedings{HgVT2025,
    author    = {Fixelle, Joshua},
    title     = {Hypergraph Vision Transformers: Images are More than Nodes, More than Edges},
    booktitle = {2025 IEEE/CVF Conference on Computer Vision and Pattern Recognition (CVPR)},
    year      = {2025},
    pages     = {9751-9761}
}

@inproceedings{HSL2022,
  title     = {Hypergraph Structure Learning for Hypergraph Neural Networks},
  author    = {Cai, Derun and Song, Moxian and Sun, Chenxi and $et$ $al$},
  booktitle = {Proceedings of the Thirty-First International Joint Conference on
               Artificial Intelligence, {IJCAI-22}},
  pages     = {1923-1929},
  year      = {2022},
}

@INPROCEEDINGS{MVTec,
  author={Bergmann, Paul and Fauser, Michael and Sattlegger, David and Steger, Carsten},
  booktitle={2019 IEEE/CVF Conference on Computer Vision and Pattern Recognition (CVPR)}, 
  title={{MVTec AD} - A Comprehensive Real-World Dataset for Unsupervised Anomaly Detection}, 
  year={2019},
  volume={},
  number={},
  pages={9584-9592},
}

@InProceedings{VisA,
    author="Zou, Yang
    and Jeong, Jongheon
    and Pemula, Latha
    and Zhang, Dongqing
    and Dabeer, Onkar",
    title={{SPot}-the-Difference Self-supervised Pre-training for Anomaly Detection and Segmentation},
    booktitle = {European Conference on Computer Vision (ECCV)},
    year="2022",
    address="Cham",
    pages="392-408",
}

@INPROCEEDINGS{BTAD,
  author={Mishra, Pankaj and Verk, Riccardo and Fornasier, Daniele and $et$ $al$},
  booktitle={2021 IEEE 30th International Symposium on Industrial Electronics (ISIE)}, 
  title={{VT-ADL}: A Vision Transformer Network for Image Anomaly Detection and Localization}, 
  year={2021},
  volume={},
  number={},
  pages={01-06},
}

@INPROCEEDINGS{MPDD,
  author={Jezek, Stepan and Jonak, Martin and Burget, Radim and Dvorak, Pavel and Skotak, Milos},
  booktitle={2021 13th International Congress on Ultra Modern Telecommunications and Control Systems and Workshops (ICUMT)}, 
  title={Deep learning-based defect detection of metal parts: evaluating current methods in complex conditions}, 
  year={2021},
  volume={},
  number={},
  pages={66-71},
}

@INPROCEEDINGS{BrainMRI,
  author={Salehi, Mohammadreza and Sadjadi, Niousha and Baselizadeh, Soroosh and Rohban, Mohammad H. and Rabiee, Hamid R.},
  booktitle={2021 IEEE/CVF Conference on Computer Vision and Pattern Recognition (CVPR)}, 
  title={Multiresolution Knowledge Distillation for Anomaly Detection}, 
  year={2021},
  volume={},
  number={},
  pages={14897-14907},
}

@INPROCEEDINGS{LiverAD,
  author = {Bao, Jinan and Sun, Hanshi and Deng, Hanqiu and He, Yinsheng and Zhang, Zhaoxiang and Li, Xingyu},
  booktitle = {2024 IEEE/CVF Conference on Computer Vision and Pattern Recognition Workshops (CVPRW)}, 
  title={{BMAD}: Benchmarks for Medical Anomaly Detection}, 
  year={2024},
  volume={},
  number={},
  pages={4042-4053},
}

@article{BUSI,
  title={Dataset of breast ultrasound images},
  author={Al-Dhabyani, Walid and Gomaa, Mohammed and Khaled, Hussien and Fahmy, Aly},
  journal={Data in Brief},
  volume={28},
  pages={104863},
  year={2020},
}

@misc{OpenCLIP,
  author       = {Ilharco, Gabriel and
                  Wortsman, Mitchell and
                  Wightman, Ross and
                  $et$ $al$},
  title        = {{OpenCLIP}},
  year         = 2021,
  version      = {0.1},
}

@inproceedings{IIPAD,
    title={One-for-all Few-Shot Anomaly Detection via Instance-Induced Prompt Learning},
    author={Wenxi Lv and Qinliang Su and Wenchao Xu},
    booktitle={International Conference on Learning Representations (ICLR)},
    year={2025},
}

@ARTICLE{Focal2020,
  author={Lin, Tsung-Yi and Goyal, Priya and Girshick, Ross and He, Kaiming and Dollár, Piotr},
  journal={IEEE Transactions on Pattern Analysis and Machine Intelligence (TPAMI)}, 
  title={Focal Loss for Dense Object Detection}, 
  year={2020},
  volume={42},
  number={2},
  pages={318-327},
}

@INPROCEEDINGS{Dice2016,
  author={Milletari, Fausto and Navab, Nassir and Ahmadi, Seyed-Ahmad},
  booktitle={2016 Fourth International Conference on 3D Vision (3DV)}, 
  title={{V-Net}: Fully Convolutional Neural Networks for Volumetric Medical Image Segmentation}, 
  year={2016},
  volume={},
  number={},
  pages={565-571},
}

@inproceedings{OneNIP2024,
  title={Learning to Detect Multi-class Anomalies with Just One Normal Image Prompt},
  author={Gao, Bin-Bin},
  booktitle = {European Conference on Computer Vision (ECCV)},
  pages={-},
  year={2024}
}

\end{document}